\documentclass[authoryear,5p,times]{elsarticle}
\RequirePackage{amsmath}
\usepackage{graphicx} 
\usepackage{pgfplots}
\pgfplotsset{compat=1.11}

\usepackage{hyperref}
\usepackage{comment}
\usepackage{amssymb}
\usepackage[utf8]{inputenc}
\usepackage{textcomp}
\usepackage{siunitx}
\usepackage{adjustbox}
\usepackage{mathtools,lmodern,tikz}
\usetikzlibrary{bending}
\usetikzlibrary{arrows}
\usepackage{verbatim}
\usepackage{caption}
\usepackage{subcaption}
\usepackage{tabularx,ragged2e,booktabs,tabulary}
\usepackage{array}
\usepackage{tkz-graph}
\usepackage{tkz-euclide}
\usepackage{arydshln}
\usepackage{natbib} %bibliography

\usepackage{colortbl,hhline}% for tables
\usepackage{tikz}
\usetikzlibrary{positioning}
\usepackage{rubikcube} 
\usepackage{url}

\usepackage{subfloat}

\definecolor{lavander}{cmyk}{0,0.48,0,0}
\definecolor{violet}{cmyk}{0.79,0.88,0,0}
\definecolor{burntorange}{cmyk}{0,0.52,1,0}

\def\lav{lavander!90}
\def\oran{orange!30}

\tikzstyle{peers}=[draw,circle,violet,bottom color=\lav,
                  top color= white, text=violet,minimum width=10pt]
\tikzstyle{superpeers}=[draw,circle,burntorange, left color=\oran, text=violet,minimum width=20pt]
\tikzstyle{legendsp}=[rectangle, draw, burntorange, rounded corners,thin,bottom color=\oran, top color=white,            text=burntorange, minimum width=2.5cm]
\tikzstyle{legendp}=[rectangle, draw, violet, rounded corners, thin,bottom color=\lav, top color= white,                     text= violet, minimum width= 2.5cm]
\tikzstyle{legend_general}=[rectangle, rounded corners, thin,
                           burntorange, fill= white, draw, text=violet,
                           minimum width=2.5cm, minimum height=0.8cm]

\usepackage{listings}
\usepackage{color}
\definecolor{red}{rgb}{0.6,0,0} 
\definecolor{blue}{rgb}{0,0,0.6}
\definecolor{green}{rgb}{0,0.8,0}
\definecolor{cyan}{rgb}{0.0,0.6,0.6}

\newcolumntype{P}[1]{>{\centering\arraybackslash}p{#1}}

\journal{Arxiv}

\begin{document}

\begin{frontmatter}

%\title{Spatial reasoning about object rotations: solving the cube comparison test in a video game}  

\title{A Qualitative Model to Reason about Object Rotations -- applied to solve the Cube Comparison Test}
 \author[label1,cor1]{Zoe Falomir}
  \ead{zfalomir@cs.umu.se}
  \cortext[cor1]{Corresponding author.}    
 \address[label1]{Umeå University, Computing Science Department, Sweden} 
\date{}

                             % keyword list or with the 

\begin{abstract}                          % Abstract of not more than 200 words.
This paper presents a Qualitative model for Reasoning about Object Rotations ($QOR$) which is applied to solve the Cube Comparison Test (CCT) by \cite{Ekstrom76}. A conceptual neighborhood graph relating the Rotation movement to the Location change and the  Orientation change ($CNG_{RLO}$) of the features on the cube sides has been built and it produces composition tables to calculate inferences for reasoning about rotations.

%Moreover, a videogame called \textsl{Treasure Hunt by Rotation}  has been developed to implement and test the $QOR$.
%The $QOR$ reasoning capability allows this videogame: (i) to provide feedback to players when they answer wrongly and (ii) to create automatically random question-answer sets used in the Mastermode. The \textsl{Treasure Hunt by Rotation} has been implemented using Unity engine and it is available to download from GooglePlay and AppleStore for everyone to train their spatial reasoning skills. Based on the CCT, a dataset of Chest Comparison Tests has been created for the test-part of the videogame and it is also provided in this paper.

\end{abstract}

\end{frontmatter}

%%%%%%%%%%%%%%%%%%%%%%%%%%%
%
% Introduction
%
%%%%%%%%%%%%%%%%%%%%%%%%%%%
\section{Introduction}
\label{sec:intro}
Studies in the literature  \citep{Wai_et_al:2009}  show that spatial reasoning skills correlate with success in Science, Technology, Engineering and Math (STEM) disciplines. Moreover, 
 spatial ability has a unique role in the development of creativity or creative-thinking (measured by patents and publications) \citep{Kell-2013}.

Spatial reasoning skills are fundamental: in medicine, for visualizing the result of a surgery; in chemistry, for understanding the structure of molecules; in engineering for designing and manufacturing 3D objects (i.e. bridges, aircrafts); in education and science communication, when reprinting visuospatial information in charts, maps, diagrams, etc.
Spatial reasoning is not an innate ability, since it has been shown that it can be trained \citep{Sorby:2009} and showed a lasting performance \citep{Uttal-2013}.
For this reason, researchers study the actualities of training spatial reasoning: in contemporary school mathematics \citep{Sinclair-Bruce:2014}, in engineering graphic courses at university \citep{Sorby:2009}, in geoscience courses at university \citep{Gold-2018}, etc.

Previous works by Falomir \textit{et al.} showed that qualitative models are useful to represent knowledge and to reason in order to solve spatial reasoning tests \citep{Falomir-LQMR:2015,mta/FalomirTPG21}. 
A qualitative descriptor for solving paper folding tests was defined by establishing a correspondence between the possible folding actions and the areas in the paper where a hole can be punched; this descriptor was tested in a videogame developed to train spatial reasoning skills on users' \citep{mta/FalomirTPG21}. Moreover, a qualitative descriptor for reasoning about 3D perspectives was developed and tested in Prolog \citep{Falomir-LQMR:2015}; then a videogame was also developed for users' training \citep{intenv/FalomirO16}.
In this paper, a new qualitative descriptor for reasoning about object rotations (QOR) is presented and implemented in to reason about how object rotations change the localisation and orientation of their sides ant to present an interactive version of the Cube Comparison Test.

In the literature, Qualitative Spatial Representations and Reasoning (QSR)  \citep{Cohn_Renz:2007,Ligozat_2011}  models and reasons about properties of \textit{space} (i.e. topology, location, direction, proximity, geometry, intersection, etc.) and their evolution between continuous neighbouring situations. QSR models have been applied to AI, as an example, qualitative descriptors of shape, colour, location and topology were used to extract logics from images ($QIDL^{+}$) and applied in ambient intelligence \citep{NeurocomFalomir:2015,jaise/Falomir17} and robotics \citep{Zfalomir_JSCC:2011,Falomir_PRL:2013}. In cognitive science, qualitative models  \citep{Forbus:2011} have also been successful to solve perceptual tests in object sketch recognition \citep{Lovett}, oddity tasks \citep{Lovett-Cognition2011}, 
%3D perspective description matching \citep{Falomir-LQMR:2015}, paper folding reasoning \citep{Falomir-QR:2016} 
and Raven’s Progressive Matrices  \citep{Lovett-Forbus:2017}. As far as we are concern, QSR have been never applied to solve the cube comparison test.

The research questions that this paper addresses are the following: 

\textit{How can we model rotation movements when manipulating a 3D object? Which is the relation between object sides? 
And which relation does exist between the rotation of an object and the location and orientation of its sides?}

\textit{Can an artificial agent solve a cube comparison question? Which reasoning mechanism does this agent need? Can this reasoning mechanism be automated and be explainable to humans?} 

This paper presents a model for reasoning about 3D object rotations, the Qualitative Object Rotation ($QOR$) which answers the  previous research questions.

The rest of the paper is organised as follows. Section \ref{sec:cube-test} presents the Cube Comparison Test. Section  \ref{sec:QOR} presents the Qualitative model for Object Rotations (QOR). 
Section \ref{sec:heuristic} outlines an algorithm to solve the Cube Comparison Test using the QOR model.
%Section \ref{sec:QOR-game}  describes the \textsl{Treasure Hunt by Rotation} videogame and its implementation. 
%Finally, conclusions and future work are presented.

%%%%%%%%%%%%%%%%%%%%%%%%%%%%%%%%%%%%%%%%%%%%%%%%%%%
\section{The Cube Comparison Test (CCT)}
\label{sec:cube-test}

The Cube Comparison Test (CCT) was developed by \cite{Ekstrom76} and it is a test of 3 minutes of duration and 21 items where participants are asked to decide if two cubes can be the same when viewed from different perspectives (no side is repeated on the same object). 
If the two cubes could be the same, participants should mark ``s''; if they could not be the same cube, participants should mark ``d" for different. 

%%%%%%%%%%%%%%%%%%%%%%%%%%%%%%%%%%%%%%%%%%%%%%
%%%%%%%%%%%%%%%%%%%%%%%%%%%%%%%%%%%%%%%%%%%%%%
\begin{figure}[h!]
\begin{minipage}{8.5cm}
\footnotesize{\textit{Wooden blocks such as children play with are often cubical with a different letter, number or symbol on each of the six faces (top, bottom, four sides). Each problem in this test consists of drawings of pairs of cubes or blocks of this kind. Remember, there is a different design, number, or letter on each face of a given cube or block. Compare the two cubes in each pair below.}}
\end{minipage}

\vspace{+0.25cm}
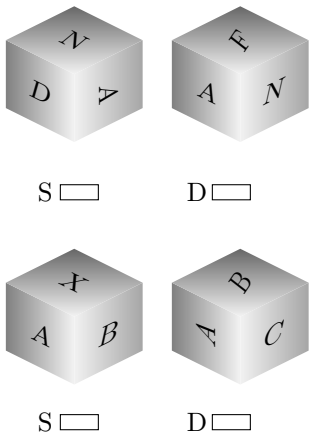
\begin{minipage}{1.25cm}
\begin{tikzpicture}[every node/.style={minimum size=0.5cm},on grid, scale=0.3]
\begin{scope}[every node/.append style={yslant=-0.5},yslant=-0.5]
  \shade[right color=gray!10, left color=black!50] (0,0) rectangle +(3,3);
  \node at (1.5,1.5) {$D$};
  %\draw (0,0) grid (3,3);
\end{scope}
\begin{scope}[every node/.append style={yslant=0.5},yslant=0.5]
  \shade[right color=gray!70,left color=gray!10] (3,-3) rectangle +(3,3);
  \node[rotate=-90] at (4.5,-1.5) {$A$};
  %\draw (3,-3) grid (6,0);
\end{scope}
\begin{scope}[every node/.append style={yslant=-0.5,xslant=0.5},yslant=0.5,xslant=-1]
  \shade[bottom color=gray!10, top color=black!60] (6,3) rectangle +(-3,-3);
  \node at (4.5,1.5) {$N$};
  %\draw (3,0) grid (6,3);
\end{scope}
\end{tikzpicture}
\end{minipage}
\hspace{+0.7cm}
\begin{minipage}{1.25cm}
\begin{tikzpicture}[every node/.style={minimum size=0.5cm},on grid, scale=0.3]
\begin{scope}[every node/.append style={yslant=-0.5},yslant=-0.5]
  \shade[right color=gray!10, left color=black!50] (0,0) rectangle +(3,3);
  \node at (1.5,1.5) {$A$};
  %\draw (0,0) grid (3,3);
\end{scope}
\begin{scope}[every node/.append style={yslant=0.5},yslant=0.5]
  \shade[right color=gray!70,left color=gray!10] (3,-3) rectangle +(3,3);
  \node at (4.5,-1.5) {$N$};
  %\draw (3,-3) grid (6,0);
\end{scope}
\begin{scope}[every node/.append style={yslant=-0.5,xslant=0.5},yslant=0.5,xslant=-1]
  \shade[bottom color=gray!10, top color=black!60] (6,3) rectangle +(-3,-3);
  \node[rotate=+90] at (4.5,1.5) {$F$};
  %\draw (3,0) grid (6,3);
\end{scope}
\end{tikzpicture}
\end{minipage}
\hspace{+0.7cm}
\begin{minipage}{4cm}
\footnotesize{\textit{The first pair have different cubes. If the left cube is turned so that the A is upright and facing you. The N would be to the left of the A and hidden, not to the right of the A as is shown on the right hand member of the pair.}}
\end{minipage}

\begin{minipage}{4cm}
\begin{tikzpicture}
\node at (0,.1) { };
\node at (0.4,.1) {S};
\node at (2.4,.1) {D};
\draw(0.6,.2)--(1.1,.2)--(1.1,0)--(0.6,0)--cycle;
\draw(2.6,.2)--(3.1,.2)--(3.1,0)--(2.6,0)--cycle;
\end{tikzpicture}
\end{minipage}

\vspace{+0.25cm}
\begin{minipage}{1.25cm}
\begin{tikzpicture}[every node/.style={minimum size=0.5cm},on grid, scale=0.3]
\begin{scope}[every node/.append style={yslant=-0.5},yslant=-0.5]
  \shade[right color=gray!10, left color=black!50] (0,0) rectangle +(3,3);
  \node at (1.5,1.5) {$A$};
  %\draw (0,0) grid (3,3);
\end{scope}
\begin{scope}[every node/.append style={yslant=0.5},yslant=0.5]
  \shade[right color=gray!70,left color=gray!10] (3,-3) rectangle +(3,3);
  \node at (4.5,-1.5) {$B$};
  %\draw (3,-3) grid (6,0);
\end{scope}
\begin{scope}[every node/.append style={yslant=-0.5,xslant=0.5},yslant=0.5,xslant=-1]
  \shade[bottom color=gray!10, top color=black!60] (6,3) rectangle +(-3,-3);
  \node at (4.5,1.5) {$X$};
  %\draw (3,0) grid (6,3);
\end{scope}
\end{tikzpicture}
\end{minipage}
\hspace{+0.7cm}
\begin{minipage}{1.25cm}
\begin{tikzpicture}[every node/.style={minimum size=0.5cm},on grid, scale=0.3]
\begin{scope}[every node/.append style={yslant=-0.5},yslant=-0.5]
  \shade[right color=gray!10, left color=black!50] (0,0) rectangle +(3,3);
  \node[rotate=90] at (1.5,1.5) {$A$};
  %\draw (0,0) grid (3,3);
\end{scope}
\begin{scope}[every node/.append style={yslant=0.5},yslant=0.5]
  \shade[right color=gray!70,left color=gray!10] (3,-3) rectangle +(3,3);
  \node at (4.5,-1.5) {$C$};
  %\draw (3,-3) grid (6,0);
\end{scope}
\begin{scope}[every node/.append style={yslant=-0.5,xslant=0.5},yslant=0.5,xslant=-1]
  \shade[bottom color=gray!10, top color=black!60] (6,3) rectangle +(-3,-3);
  \node[rotate=90] at (4.5,1.5) {$B$};
  %\draw (3,0) grid (6,3);

\end{scope}
\end{tikzpicture}
\end{minipage}
\hspace{+0.7cm}
\begin{minipage}{4cm}
\footnotesize{\textit{The 2nd pair correspond to different views of the same cube. Note that, if the A is turned on its side the X becomes hidden, the B is now on top, and the C (which was hidden) now appears. }}
\end{minipage}

\begin{minipage}{4cm}
\begin{tikzpicture}
\node at (0,.1) { };
\node at (0.4,.1) {S};
\node at (2.4,.1) {D};
\draw(0.6,.2)--(1.1,.2)--(1.1,0)--(0.6,0)--cycle;
\draw(2.6,.2)--(3.1,.2)--(3.1,0)--(2.6,0)--cycle;
\end{tikzpicture}
\end{minipage}

\caption{Example question and instructions given in the Cube Comparison Test (CCT) from the \textit{Manual for Kit of Factor-Referenced Cognitive Tests} by \cite{Ekstrom76}.}
\label{fig:example-question}
\end{figure}
%%%%%%%%%%%%%%%%%%%%%%%%%%%%%%%%%%%%%%%%%%%%%%%%%%%%%%%%%%%%%%%
%%%%%%%%%%%%%%%%%%%%%%%%%%%%%%%%%%%%%%%%%%%%%%%%%%%%%%%%%%%%%%%

The CCT has also been used extendedly in evaluating spatial reasoning skills, for example: (i) by the Educational Testing Service in Princeton, New Jersey \citep{thomasr.lordjoyl.rupert1995}, (ii) as one part of the 11+ exam for students in England and Northern Ireland\footnote{Eleven plus Exams Head for success: \href{https://www.elevenplusexams.co.uk}{examples} (Accessed June 2018)}; (iii) to evaluate surgical trainees' visual spatial ability which plays a role in fast learning endoscopic and laparoscopic surgery \citep{Henn2018}, etc.

The CCT has also been used in functional Magnetic Resonance Imaging (fMRI) for studying which parts of the human brain are involved in mental rotation (e.g. in the studies by \citep{WINDISCHBERGER:2003,LAMM2001}). For that, the \textit{Dreidimensionaler W{\"u}rfeltest (3 DW)} stimuli by \cite{gittler1990dreidimensionaler} were used. Recently, a variation of the 3DW %\citep{gittler1990dreidimensionaler}
has appeared in the literature: the R-Cube-SR Test by \cite{Fehringer:2023}, where participants do not need to check the orientation of the symbols on the sides of the cube (i.e. letters) to complete the task. This indicates that the CCT is still of interest in multidisciplinar areas of science.

%%%%%%%%%%%%%%%%%%%%%%%%%%
%
% Model
%
%%%%%%%%%%%%%%%%%%%%%%%%%

\section{The Qualitative Model of Object Rotations}
\label{sec:QOR} 

This section presents the Qualitative Model of Object Rotation (QOR) which studies the relations between the perspectives of an object and how they change depending on the rotations applied to it.

Qualitative models are defined by a descriptor based on reference systems and operators:

\begin{center}
$QOR$ = \texttt{<} QOD$_{RS}$, QOR$_{RS}$\texttt{>}\\
\end{center}

\noindent where the QOD$_{RS}$ describes an object O (or it associated bounding box) which has 6 canonical sides parallel in pairs and $view_{O_{xyz}}$ is the view of that object O  defined by 3 sides (x,y,z) which are perpendicular to each other. Each side is characterized by a feature (content), a location and an orientation. The type of features contained by each side can vary from a simple symbol to an image/texture or a range of depths depending on the pattern recognition techniques used (i.e. pixels in a digital image, points in a RGB-Depth point cloud, etc). The  QOD$_{RS}$ is described in Section \ref{sec:QOD}  and it was first presented by \cite{mdai/FalomirC25}.

This paper presents for the first time the rotation reference system (QOR$_{RS}$), that is, the rotation movements/operations associated to our QOD representation and how they correspond to possible actions that can be applied to a QOD so that its features change their location and orientation from the point of view of the observer.

\subsection{The Qualitative Object Descriptor (QOD)}
\label{sec:QOD} 

The QOD \citep{mdai/FalomirC25} describes any three-dimensional object (or its bounding box) as:

\begin{center}
$view(Object_{ID}, Perspective_{x,y,z})=$

$\{QOD_x, QOD_y, QOD_z\}$
\end{center}

\noindent that is,

$view(Object_{ID}, Perspective_{x,y,z})=\{$

$QOD_x: (Feature_{RS}, Location_{RS}, Orientation_{RS}),$

$QOD_y: (Feature_{RS}, Location_{RS}, Orientation_{RS}),$

$QOD_z: (Feature_{RS}, Location_{RS}, Orientation_{RS})\}$\\

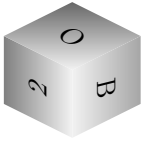
\begin{figure}[h!]
\begin{minipage}{1cm}
\begin{tikzpicture}[every node/.style={minimum size=0.5cm},on grid, scale=0.3]
\begin{scope}[every node/.append style={yslant=-0.5},yslant=-0.5]
  \shade[right color=gray!10, left color=black!50] (0,0) rectangle +(3,3);
   \node[rotate=+90] at (1.5,1.5) {$2$};
  %\draw (0,0) grid (3,3);
\end{scope}
\begin{scope}[every node/.append style={yslant=0.5},yslant=0.5]
  \shade[right color=gray!70,left color=gray!10] (3,-3) rectangle +(3,3);
  \node[rotate=-90]  at (4.5,-1.5) {$B$};
  %\draw (3,-3) grid (6,0);
\end{scope}
\begin{scope}[every node/.append style={yslant=-0.5,xslant=0.5},yslant=0.5,xslant=-1]
  \shade[bottom color=gray!10, top color=black!60] (6,3) rectangle +(-3,-3);
  \node at (4.5,1.5) {$O$};
  %\draw (3,0) grid (6,3);
\end{scope}
\end{tikzpicture}
\end{minipage}
\hspace{+1.7cm}
\begin{minipage}{4cm}
$View(obj_1,t_1)= \{$

$(``G", front, 3q),$

$(``B", right, 1q),$

$(``O", up, non$-$oriented)\}$
\end{minipage}
\caption{Example of an object view described by the QOR model.}
\label{fig:vO}
\end{figure}

\figurename{} \ref{fig:vO} shows an
example of an object containing a symbol ``A" on its front side (no-turned), containing a symbol ``B" on its right side (no-turned) and containing a symbol ``X" on its up side (non-oriented, since it is a symmetric symbol).

%%%%%%%%%%%%%%%%%%%%%%%%%%%%%%%%%%%%%%%%%%%%%%
%%%%%%%%%%%%%%%%%%%%%%%%%%%%%%%%%%%%%%%%%%%%%%%%%%%%%%%%%%%%%%%
%%%%%%%%%%%%%%%%%%%%%%%%%%%%%%%%%%%%%%%%%%%%%%%%%%%%%%%%%%%%%%%

Each side of each object contains a feature which is spatially described by its location and its orientation. 
Thus, each object dimension is described by the following reference systems (RS):
$$QO_{side \in \{x,y,z\}}= \{Feature_{RS}, Location_{RS},Orientation_{RS}\} $$

The $Feature_{RS}$ describes the object side features as: 

$$Feature_{RS} = \{Feature_{D}, Feature_{N}\}$$ 

\noindent where $Feature_{D}$ is the descriptor of the feature (e.g. a set of pixels corresponding to a symbol, voxels, etc.) which has been grounded by pattern recognition to a symbol/concept/name or $Feature_{N}$. In the examples in this paper, we use a set of caption letters from the latin alphabet and a set of numbers as $Feature_{D}$. In the videogame application shown in the experimentation section, the features in $Feature_{D}$ are as set of textures that show object drawings. \\

The Location Reference System describes in which side of the object is each feature situated:
$$Location_{RS} = \{(x,y,z), Location_{N}, Location_{G}\}$$ 

\noindent Location$_{N}$ $\in$  \{$front(f)$, $right(r)$, $up(u)$, $down(d)$, $back (b)$, $left (f)$\}

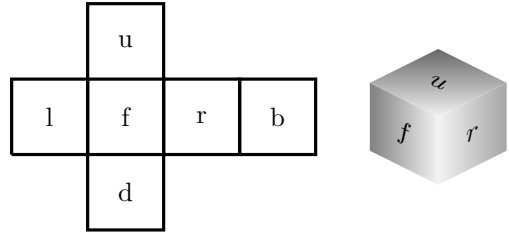
\begin{figure}[h!]
\begin{minipage}{2cm}
\begin{tikzpicture}[scale=.25]
  \begin{scope}
    \draw[very thick, scale=1](0,0)--(4,0)--(4,4)--(0,4)--(0,0);
    \node[anchor=center] at (2, 2) {l};    
   \draw[very thick, scale=1](4,0)--(8,0)--(8,4)--(4,4)--(4,0);
    \node[anchor=center] at (6, 2) {f};    
    \draw[very thick, scale=1](4,4)--(4,8)--(8,8)--(8,4);
    \node[anchor=center] at (6, 6) {u};   
    \draw[very thick, scale=1](4,0)--(4,-4)--(8,-4)--(8,0);
    \node[anchor=center] at (6, -2) {d};   
    \draw[very thick, scale=1](8,0)--(12,0)--(12,4)--(8,4)--(8,0);
    \node[anchor=center] at (10, 2) {r};
    \draw[very thick, scale=1](12,0)--(16,0)--(16,4)--(12,4)--(12,0);
    \node[anchor=center] at (14, 2) {b};
  \end{scope}
\end{tikzpicture}
\end{minipage}
\hspace{+2.5cm}
\begin{minipage}{1cm}
\begin{tikzpicture}[every node/.style={minimum size=0.5cm},on grid, scale=0.3]
\begin{scope}[every node/.append style={yslant=-0.5},yslant=-0.5]
  \shade[right color=gray!10, left color=black!50] (0,0) rectangle +(3,3);
  \node at (1.5,1.5) {$f$};
  %\draw (0,0) grid (3,3);
\end{scope}
\begin{scope}[every node/.append style={yslant=0.5},yslant=0.5]
  \shade[right color=gray!70,left color=gray!10] (3,-3) rectangle +(3,3);
  \node at (4.5,-1.5) {$r$};
  %\draw (3,-3) grid (6,0);
\end{scope}
\begin{scope}[every node/.append style={yslant=-0.5,xslant=0.5},yslant=0.5,xslant=-1]
  \shade[bottom color=gray!10, top color=black!60] (6,3) rectangle +(-3,-3);
  \node at (4.5,1.5) {$u$};
  %\draw (3,0) grid (6,3);
\end{scope}
\end{tikzpicture}
\end{minipage}
\caption{Perspective front-right-up.}
\label{fig:FRU}
\end{figure}

\noindent Location$_{G}$ $\in$  \{$u \bot f$, $ u \bot b$, $ u \bot r$, $u \bot l$, $d \bot f$, $d \bot b$, $d \bot r$, $d \bot l$, $f\parallel b$, $u \parallel d$, $r \parallel l$ \}  \\

\noindent where $(x,y,z)$ indicate coordinates in space,
Location$_{N}$ presents the locations of the sides on the cube with respect to the point of view of the observer,
Location$_{G}$ describes the geometric representation of these locations in the 3D space corresponding to each previously defined concept, respectively.
Note that each side has 4 neighbouring sides and one opposite side. The four neighbouring sides are located geometrically perpendicular to the original side and are parallel in pairs. The opposite sides are parallel to each other.
\figurename{} \ref{fig:FRU} shows a cube and how the object sides are unfolded taking as reference the \textit{front} side (f). In the unfolding drawing it is straightforward to recognise that the \textit{front}-side neighbours are \textit{left}, \textit{right}, \textit{up} and \textit{down} sides (which are located on perpendicular planes), whereas \textit{back} is its opposite side (located on a parallel plane).

%%%%%%%%%%%%%%%%%%%%%%%%%%%%%%%%%%%%%%%%%%%%%%%
%%%%%%%%%%%%%%%%%%%%%%%%%%%%%%%%%%%%%%%%%%%%%%%%%%%%%

The orientation of the features in the object is described according to an Orientation Reference System or 
$Orientation_{RS}$ which has the following components:\\

Orientation$_{A}$ $\in$ \{1q, 2q,  3q, 0q, \textit{none-oriented}\}

Orientation$_{G_1}$ $\in$ \{ $90^\circ$, $180^\circ$, $270^\circ$, $0$, any\}\\

Orientation$_{R}$ $\in$ \{+q, +2q,  -2q, -q, \textit{same}\}

Orientation$_{G_2}$ $\in$ \{ $+90^\circ$, $+180^\circ$, $-180^\circ$, $-90^\circ$, $0$\}\\

\noindent where degrees ($^\circ$) indicate the unit of measurement of the orientation; Orientation$_{A}$ refers to the set of concepts or names that define a specific orientation, e.g. ``turned a quarter clockwise (1q)"; and Orientation$_{G_1}$ refers to the geometric counterpart, that is, the turning angle clockwise in degrees ($^\circ$) which is incrementing in steps of  ($90^\circ$), Orientation$_{R}$ refers to the set of concepts that define relative orientation, e.g. ``orientation increased a quarter (+q)" and Orientation$_{G_1}$ define to the corresponding relative turning angles.

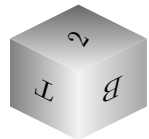
\begin{figure}[h!]
\begin{minipage}{1cm}
\begin{tikzpicture}[every node/.style={minimum size=0.5cm},on grid, scale=0.3]
\begin{scope}[every node/.append style={yslant=-0.5},yslant=-0.5]
  \shade[right color=gray!10, left color=black!50] (0,0) rectangle +(3,3);
   \node[rotate=-180] at (1.5,1.5) {$T$};
\end{scope}
\begin{scope}[every node/.append style={yslant=0.5},yslant=0.5]
  \shade[right color=gray!70,left color=gray!10] (3,-3) rectangle +(3,3);
  \node[rotate=-180]  at (4.5,-1.5) {$B$};
\end{scope}
\begin{scope}[every node/.append style={yslant=-0.5,xslant=0.5},yslant=0.5,xslant=-1]
  \shade[bottom color=gray!10, top color=black!60] (6,3) rectangle +(-3,-3);
  \node[rotate=90]  at (4.5,1.5) {$2$};
\end{scope}
\end{tikzpicture}
\end{minipage}
\hspace{+1.7cm}
\begin{minipage}{4cm}
$View(obj_1, t_2)= \{$

$(``T", front, 2q),$

$(``B", right, 2q),$

$(``G", up, 3q)\}$
\end{minipage}
\caption{Example of the same cube view in \figurename{} \ref{fig:vO} but rotated, discovering a new feature (T) and changing the orientations of two common features (G and B).}
\label{fig:vO2}
\end{figure}

A cube side can be turned one-quarter (1q or $90^\circ$), two-quarters (2q or $180^\circ$), three-quarters (3q or $270^\circ$ or $-90^\circ$), or not being turned at all.
Note the symbols in the cube shown in \figurename{} \ref{fig:vO2} the orientation of feature ``B" has increased a quarter (+q) with respect to it orientation in the cube shown in \figurename{} \ref{fig:vO}, that is, evolves from orientation 1q to 2q. Note also that the orientation of feature ``G" stays the same, that is, it stays oriented three quarters or 3q. Finally note that there is a new feature appearing in \figurename{} \ref{fig:vO2}, feature ``T" which we can describe as being upside-down or turned 2-quarters with respect to its conventional use in linguistics.

%%%%%%%%%%%%%%%%%%%%%%%%%%%%%%%%%%%%%%%%%%%%%%
%%%%%%%%%%%%%%%%%%%%%%%%%%%%%%%%%%%%%%%%%%%%%%%%%%%%%%%%%%%%%%%

%%%%%%%%%%%%%%%%%%%%%%%%%%%
%
% Rotations
%
%%%%%%%%%%%%%%%%%%%%%%%%%%%%

%%%%%%%%%%%%%%%%%%%%%%%%%%%%%%%%%%%%%%%%
\subsection{Rotations in QODs (QOR$_{RS}$) }
\label{sec:OP}
The operators associated to a representation correspond to the possible actions that can be applied to an object so that it changes its features.
Any rotation on an object changes the location of the features on its sides and also how these features are oriented.

This section describes the possible rotation operators (QOR$_{RS}$) applied to an object and its corresponding geometric counterparts.
Thus, the operators associated to the QOR are defined by the following Reference System as: \\

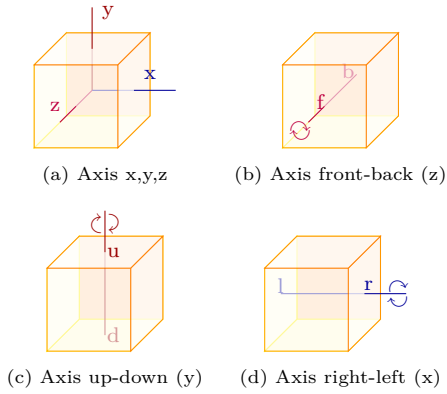
\begin{figure}[ht] \centering
\caption{Operators for QOR: rotating objects depending on 3 axes in two possible directions. }
\label{fig:simplerotations}
\begin{adjustbox}{center}
\newcommand{\Depth}{2}
\newcommand{\Height}{2}
\newcommand{\Width}{2}

\begin{minipage}{10cm}
\footnotesize
\centering
\begin{subfigure}[b]{0.3\textwidth }
\centering
\label{fig:simplerotationsSub1}

\begin{tikzpicture}[scale=0.55]
\coordinate (O) at (0,0,0);
\coordinate (A) at (0,\Width,0);
\coordinate (B) at (0,\Width,\Height);
\coordinate (C) at (0,0,\Height);
\coordinate (D) at (\Depth,0,0);
\coordinate (E) at (\Depth,\Width,0);
\coordinate (F) at (\Depth,\Width,\Height);
\coordinate (G) at (\Depth,0,\Height);

%Linea azul trasera para centrar el cubo
%\draw[blue] (\Width,\Height/2,\Depth/2) -- (\Width*2,\Height/2,\Depth/2);

\draw[yellow,fill=yellow!20] (O) -- (C) -- (G) -- (D) ;% Bottom side
\draw[yellow,fill=orange!30] (O) -- (A) -- (E) -- (D) -- cycle;% Back side
\draw[yellow,fill=yellow!10] (O) -- (A) -- (B) -- (C) -- cycle;% Left side

\draw[red] (\Width/2,1,\Depth/2) -- (\Width/2,2,\Depth/2);
\draw[blue] (1,\Height/2,\Depth/2) -- (2,\Height/2,\Depth/2);
\draw[purple] (\Width/2,\Height/2,1) -- (\Width/2,\Height/2,2);
\draw[orange,fill=white!20,opacity=0.8] (D) -- (E) -- (F) -- (G) -- cycle;% Right side
\draw[orange,fill=white!20,opacity=0.6] (C) -- (B) -- (F) -- (G) -- cycle;% Front side
\draw[orange,fill=white!20,opacity=0.8] (A) -- (B) -- (F) -- (E) -- cycle;% Top side

\node[red] at (1,2.5) {y};
\node[blue] at (2,1) {x};
\node[purple] at (.5,1,2) {z};

\draw[red] (\Width/2,2,\Depth/2) -- (\Width/2,3,\Depth/2);
\draw[blue] (2,\Height/2,\Depth/2) -- (3,\Height/2,\Depth/2);
\draw[purple] (\Width/2,\Height/2,2) -- (\Width/2,\Height/2,3);

\end{tikzpicture} 
\caption{Axis x,y,z}
\end{subfigure}
\begin{subfigure}[b]{0.3\textwidth }
\centering
\label{fig:simplerotationsSub2}
\begin{tikzpicture}[scale=0.55]
%Linea azul trasera para centrar el cubo
%\draw[blue] (\Width,\Height/2,\Depth/2) -- (\Width*2,\Height/2,\Depth/2);

\draw[yellow,fill=yellow!20] (O) -- (C) -- (G) -- (D) ;% Bottom side
\draw[yellow,fill=orange!30] (O) -- (A) -- (E) -- (D) -- cycle;% Back side
\draw[yellow,fill=yellow!10] (O) -- (A) -- (B) -- (C) -- cycle;% Left side
\draw[orange,fill=white!20,opacity=0.8] (D) -- (E) -- (F) -- (G) -- cycle;% Right side
\draw[purple] (\Width/2,\Height/2,0) -- (\Width/2,\Height/2,\Depth);
\node[purple] at (0.8,1.1,0) {b};
\draw[orange,fill=white!20,opacity=0.6] (C) -- (B) -- (F) -- (G) -- cycle;% Front side
\draw[orange,fill=white!20,opacity=0.8] (A) -- (B) -- (F) -- (E) -- cycle;% Top side
\draw[purple] (\Width/2,\Height/2,\Depth+1) -- (\Width/2,\Height/2,\Depth);

\node[purple] at (.9,1.1,2) {f};
\node[purple] at (1,1.1,3.5) {$\curvearrowright$};
\node[purple] at (1,0.9,3.5) {$\mathrel{\reflectbox{\rotatebox[origin=c]{180}{$\curvearrowleft$}}}$};
%\draw[dashed] (1,1,2) -- (1,0,2);
%\draw[dashed] (1,0,2) -- (1,0,0);
%\draw[dashed] (1,0,0) -- (1,1,0);

\end{tikzpicture}
\caption{Axis front-back (z)}
\end{subfigure}
%go to the next row

\begin{subfigure}[b]{0.3\textwidth }
\centering
\label{fig:simplerotationsSub3}
\begin{tikzpicture}[scale=0.55]

%Linea azul trasera para centrar el cubo
%\draw[blue] (\Width,\Height/2,\Depth/2) -- (\Width*2,\Height/2,\Depth/2);

\draw[yellow,fill=yellow!20] (O) -- (C) -- (G) -- (D) ;% Bottom side
\draw[yellow,fill=orange!30] (O) -- (A) -- (E) -- (D) -- cycle;% Back side
\draw[yellow,fill=yellow!10] (O) -- (A) -- (B) -- (C) -- cycle;% Left side
\node[red] at (1.2,0,1) {d};
\draw[red] (1,0,1) -- (1,2,1);
\draw[orange,fill=white!20,opacity=0.8] (D) -- (E) -- (F) -- (G) -- cycle;% Right side
\draw[orange,fill=white!20,opacity=0.6] (C) -- (B) -- (F) -- (G) -- cycle;% Front side
\draw[orange,fill=white!20,opacity=0.8] (A) -- (B) -- (F) -- (E) -- cycle;% Top side

\node[red] at (1.2,2,1) {u};
\node[red] at (0.8,2.7,1) {$\mathrel{\reflectbox{\rotatebox[origin=c]{270}{$\curvearrowleft$}}}$};
\node[red] at (1.2,2.7,1) {$\mathrel{\reflectbox{\rotatebox[origin=c]{90}{$\curvearrowleft$}}}$};

\draw[red] (1,2,1) -- (1,3,1);

\end{tikzpicture}
\caption{Axis up-down (y)}
\end{subfigure}
\begin{subfigure}[b]{0.3\textwidth }
\centering
\begin{tikzpicture}[scale=0.55]

%Linea azul trasera para centrar el cubo
%\draw[blue] (\Width,\Height/2,\Depth/2) -- (\Width*2,\Height/2,\Depth/2);

\draw[yellow,fill=yellow!20] (O) -- (C) -- (G) -- (D) ;% Bottom side
\draw[yellow,fill=orange!30] (O) -- (A) -- (E) -- (D) -- cycle;% Back side
\draw[yellow,fill=yellow!10] (O) -- (A) -- (B) -- (C) -- cycle;% Left side
\node[blue] at (0,1.2,1) {l};
\draw[blue] (0,1,1) -- (2,1,1);
\draw[orange,fill=white!20,opacity=0.8] (D) -- (E) -- (F) -- (G) -- cycle;% Right side
\draw[orange,fill=white!20,opacity=0.6] (C) -- (B) -- (F) -- (G) -- cycle;% Front side
\draw[orange,fill=white!20,opacity=0.8] (A) -- (B) -- (F) -- (E) -- cycle;% Top side

\node[blue] at (2.1,1.2,1) {r};
\node[blue] at (2.8,1.2,1) {$\mathrel{\reflectbox{\rotatebox[origin=c]{-15}{$\curvearrowleft$}}}$};
\node[blue] at (2.8,0.8,1) {$\mathrel{\reflectbox{\rotatebox[origin=c]{175}{$\curvearrowleft$}}}$};

\draw[blue] (2,1,1) -- (3,1,1);

\end{tikzpicture}
\caption{Axis right-left (x)}
\end{subfigure}
\end{minipage}
\end{adjustbox}
\end{figure}

\noindent QOR$_{RS}$= \{Axis, Rotation$_{G}$, Rotation$_{N}$\} \\

\noindent where Axis is defined by the line between the centres of two opposite object sides. The cartesian axis (x,y,z) is the reference for geometric calculations: the axis \textit{x} goes from \textit{right} to \textit{left} (rl), the axis \textit{y} goes from \textit{up} to \textit{down} (ud), and the axis \textit{z}  goes from \textit{front} to \textit{back} (fb) (see \tablename{} \ref{tab:ObjectRotations} and \figurename{} \ref{fig:simplerotations} for more details\footnote{Even though more rotation axis can be found in the cube, the QOR uses the Euler definition of rotation axes on a cube which are centred on the cube centroid.}).

%%%%%%%%%%%%%%%%%%%%%%%%%%%%%%%%%
%%%%%%%%%%%%%%%%%%%%%%%%%%%%%%%%%%%%%%%%%%%%%%%

%%%%%%%%%%%%%%%%%%%%%%%%%%%%%%%%%%%%%%%%%%%%%%%%%%%%%%%%%%%%%%
\begin{comment}
%%%%%%%%
\begin{table}[ht]
\caption{The Rotation Reference System (Rotation$_{RS}$).}
\label{tab:ObjectRotations}
\footnotesize
\begin{adjustbox}{center}
\begin{tabular}{  |c|c|c|c| }\hline
 Cartesian & Qualitative Axis &\multicolumn{2}{c}{Rotations in the Axis}\\
\multicolumn{1}{c}{ Axis}&&+90\textsuperscript {o}&-90\textsuperscript {o}\\ \hline
x & right-left  &\textuparrow & \textdownarrow\\
y &up-down & \textrightarrow & \textleftarrow
\\
z &front-back  & $\curvearrowright$ & $\curvearrowleft$ \\
\hline
\end{tabular}
\end{adjustbox}
\end{table}
%
\end{comment}
%%%%%%%%%%%%%%%%%%%%%%%%%%%%%%%%%%%%%%%%%%%%%%%%%%%%%%%%%%%
%$\rotatebox[origin=c]{-135}{$\Leftarrow$}$
%$\Downarrow$

Rotation$_{G}$ refers to 90-degree rotations on the axes previously defined and  Rotation$_{N}$ refers to the qualitative names given to that rotations according to the correspondences provided in \tablename{} \ref{tab:ObjectRotations}.

%%%%%%%%%%%%%%%%%%%%%%%%%%%%%%%%%%%%%%%%%%%%
\begin{table}[h!]
\caption{The Rotation Reference System (Rotation$_{RS}$).}
\begin{adjustbox}{center}
\label{tab:ObjectRotations}
\footnotesize
\begin{tabular}{r >{\centering\arraybackslash}m{0.3in} |c|r|l}
\hline
Rotation$_G$ &  & & Rotation$_N$& \\ \hline
degrees&Cart.&Direction&Icon&Description\\ 
&Axis&&&\\ \hline
 -90&x& right-left & \textuparrow & towards-up\\ 
90&x& right-left &\textdownarrow  &towards-down \\ 
-90&y&up-down&\textleftarrow & towards-left\\ 
90&y&up-down&\textrightarrow & towards-right \\ 
-90&z& front-back & $\curvearrowright$ &towards-up-right \\ 
90&z&front-back & $\curvearrowleft$  & towards-up-left \\ 
%-90&z&fb & $\mathrel{\reflectbox{\rotatebox[origin=c]{180}{$\curvearrowleft$}}}$& towards-down-left \\ 
%90&z&fb & $\mathrel{\reflectbox{\rotatebox[origin=c]{180}{$\curvearrowright$}}}$& towards-down-right \\
%0 & any & any & - & no-movement \\ 
\hline
\end{tabular}
\end{adjustbox}
\end{table}
%%%%%%%%%%%%%%%%%%%%%%%%%%%%%%%%%%%%%%%%%%%%%%%%%%%

%Note that, the rotation $\curvearrowright$ is equivalent to $\mathrel{\reflectbox{\rotatebox[origin=c]{180}{$\curvearrowleft$}}}$, that is, they produce the same geometrical trasnformations (i.e. a -90 degree rotation in the Cartesian axis z), and that $\curvearrowleft$ is equivalent to $\mathrel{\reflectbox{\rotatebox[origin=c]{180}{$\curvearrowright$}}}$ too. \tablename{} \ref{tab:ObjectRotations} lists all possible rotation movements since cognitively, the embodied action carried out on the object is different: turning to the right or turning to the left.

%%%%%%%%%%%%%%%%%%%%%%%%%%%%%%%%%%%%%%%%%%%%%%%%%%%%
%\begin{table}[h!]
%\caption{Table of equivalent rotations of the cube}
%\label{tab:equivalent}
%\begin{adjustbox}{center}
%\begin{tabular}{| c |c | }\hline
%Original & Equivalent\\ \hline
%$\curvearrowright$ &$\mathrel{\reflectbox{\rotatebox[origin=c]{180}{$\curvearrowleft$}}}$\\ 
%$\curvearrowleft$ &$\mathrel{\reflectbox{\rotatebox[origin=c]{180}{$\curvearrowright$}}}$\\ \hline
%\end{tabular}
%\end{adjustbox}
%\end{table}
%%%%%%%%%%%%%%%%%%%%%%%%%%%%%%%%%%%%%%%%%%%%%%%%%%%%

%All rotations of the object that change the localization of the features can be expressed using the movements from \tablename{} \ref{tab:ObjectRotations}. 
%The sides of the cube are connected based on this graph \figurename{} \ref{fig:graph-rotations}, using this graph it can be seen that with one rotation from \tablename{} \ref{tab:ObjectRotations}, a side can end up being any other except the opposite one.

\subsection{Rotations related to Location and Orientation Change}
\label{sec:locChange}

\begin{figure}[h!]
\caption{Conceptual neighborhood diagram for locations ($CND_{L}$).}
\label{fig:CND}
\begin{adjustbox}{center}
\begin{tikzpicture}[scale=0.2,>=stealth',shorten >=1pt,auto,node distance=1.5cm,
                    thick,main node/.style={circle,scale=0.8,draw,font=\sffamily\Large\bfseries}]

  \node[main node] (f) {f};
  \node[main node] (r) [right of=f] {r};
  \node[main node] (d) [below  of=f] {d};
  \node[main node] (u) [above of=f] {u};
  \node[main node] (b) [right of=r] {b};
  \node[main node] (l) [left of=f] {l};
  
  \path[thick]
  (f) edge (d) (f) edge (l) (f) edge (r) (f) edge (u) (r) edge (b)
  (l)edge [bend left=45]  (u)
  (l)edge [bend left=-45]  (d)
  (u)edge [bend left=45]  (r)
  (d)edge [bend left=-45]  (r)
  (u)edge [bend left=30]  (b)
  (d)edge [bend left=-30]  (b)
  (l)edge [out=240,in=-90,looseness=1.3]  (b);
\end{tikzpicture}
\end{adjustbox}
\end{figure}
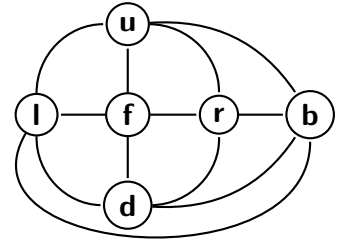

As in a cube each side has four neighbouring sides on perpendicular planes and an opposite side in a parallel plane, then each 90-degree rotation in the $Rotation_{RS}$ involves to discover only one new feature of the object. That is, after one 90-degree rotation, two features are still seen from the same perspective, although they change locations, and one new feature appears at a new location. How features change to neighbouring locations after a 90-degree rotation can be represented in a conceptual neighbourhood diagram or CND (see \figurename{} \ref{fig:CND}) where each link represents a possible transition (or rotation).

%%%%%%%%%%%%%%%%%%%%%%%%%%%%%%%%%%%%%%
%%%%%%%%%%%%%%%%%%%%%%%%%%%%%%%%%%%%%%
%%%%%%%%%%%%%%%%%%%%%%%%%%%%%%%%%%%
%%%%%%%%%%%%%%%%%%%%%%%%%%%%%%%%%%%

Note that in \figurename{} \ref{fig:CND} the nodes correspond to the locations of the sides of an object (4 neighbouring sides and one opposite side), where  each node has 4 direct transitions that connect it to their neighbours. As the four neighbouring sides are located geometrically perpendicular to the original side they can be reached by 90-degree-rotations. Note that opposite sides are parallel to each other and they cannot be reached by a 90-degree-rotation.

\begin{figure}[ht]
\caption{Conceptual neighborhood graph (CNG) relating the rotations with location changes of features: $CNG_{RL}$.}
\label{fig:graph-rotations}
\begin{adjustbox}{center}

\resizebox {\columnwidth} {!} {
\begin{tikzpicture}[->,>=stealth',shorten >=0.5pt,auto,node distance=2.8cm,
                    thick, main node/.style={circle,draw,font=\sffamily\normalsize\bfseries}]

  \node[main node] (f) {front};
  \node[main node] (r) [below right of=f] {right};
  \node[main node] (d) [below right of=r] {down};
  \node[main node] (u) [below left of=f] {up};
  \node[main node] (l) [below left of=u] {left};
  \node[main node] (b) [below right of=f][below of=u] {back};

  \path[every node/.style={font=\sffamily\small}]
    (f) edge [bend left=10] node [right] {\textrightarrow } (r)
        edge [bend left=10] node[right] {\textuparrow } (u)
        edge [bend right=115] node[left] {\textleftarrow } (l)
        edge [bend left=115] node[right] {\textdownarrow } (d)
        edge [loop above] node {$\curvearrowleft$ } (f)
        edge [loop below] node {$\curvearrowright$ } (f)
    (u) edge [bend left=5]  node  {$\curvearrowright$ } (r)
        edge [bend left=10] node[left] {\textdownarrow } (f)
        edge [bend left=10] node[right] {\textuparrow  } (b)
        edge [bend left=10] node[right] {$\curvearrowleft$ } (l)
        edge [loop above] node {\textrightarrow } (u)
        edge [loop left=10] node {\textleftarrow  } (u)
   (l)  edge[bend left=10]node  {$\curvearrowright$ } (u)
        edge [bend left=10] node[right] {\textleftarrow } (b)
        edge [bend left=60] node[right] {\textrightarrow } (f)
        edge  [bend left=10] node {$\curvearrowleft$ } (d)
        edge [loop above] node {\textdownarrow } (l)
        edge [loop below] node {\textuparrow } (l)
   (b)  edge  [bend left=10] node  {\textrightarrow } (l)
        edge [bend left=10] node {\textdownarrow } (u)
        edge [bend left=10] node {\textleftarrow } (r)
        edge [bend left=5] node {\textuparrow} (d)
       % edge [loop above] node {$\curvearrowright$ } (b)
       % edge [loop below] node {$\curvearrowleft$ } (b)
   (d)  edge [bend left=10] node  {\textdownarrow } (b)
        edge [bend left=10] node {$\curvearrowleft$ } (r)
        edge [bend left=10] node {$\curvearrowright$ } (l)
        edge [bend right=60] node[left] {\textuparrow  } (f)
        edge [loop above] node {\textrightarrow } (d)
        edge [loop below] node {\textleftarrow } (d)
    (r) edge [bend left=10] node  {\textleftarrow } (f)
        edge [bend left=10] node {\textrightarrow } (b)
        edge [bend left=10] node {$\curvearrowleft$ } (u)
        edge [bend left=10] node {$\curvearrowright$ } (d)
        edge [loop above] node {\textdownarrow } (r)
        edge [loop right] node {\textuparrow } (r) ;
        
        %aristas de B
       \path (b) edge [out=330,in=300,looseness=10] node[right] {$\curvearrowright$} (b);
       \path (b) edge [out=200,in=230,looseness=10] node[left] {$\curvearrowleft$} (b);

\end{tikzpicture}
}
\end{adjustbox}
\end{figure}
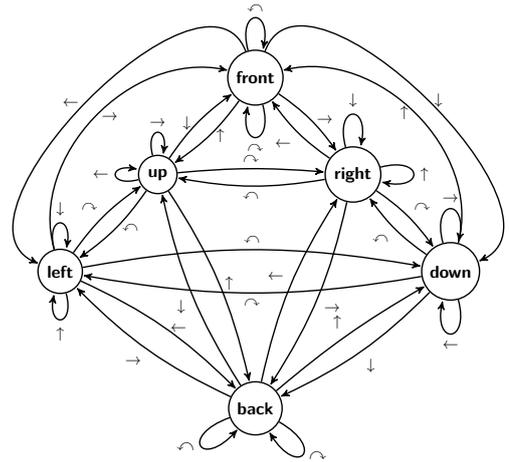

If we populate the conceptual neighbourhood diagram with the rotations in the $Rotation_{RS}$ 
then we obtain an oriented conceptual neighbour graph (CNG) of relations between the changing location of features and the rotation performed on the object: $CNG_{RL}$ (\figurename{} \ref{fig:graph-rotations}).

%%%%%%%%%%%%%%%%%%%%%%%%%%%%%%%%%%%%%%%%%%%%%
%%%%%%%%%%%%%%%%%%%%%%%%%%%%%%%%%%%%%%%%%%%%%%%%%%%

The $CNG_{RL}$ in \figurename{} \ref{fig:graph-rotations} can be used to build the inference table about location change showed in \tablename{} \  \ref{tab:matrixqualitative}.
Moreover, from \tablename{} \ref{tab:matrixqualitative}, movements which change feature locations between visible sides in the CCT (fru) can be extracted in \tablename{} \ref{tab:MostVisible} as an excerpt to solve the CCT more efficiently.

%%%%%%%%%%%%%%%%%%%%%%%%%%%%%%%%%%%%%%%%%%%%%%%%%%%%%%%%%%%%%%%%%%%%%%%%%%
\begin{table} [h!]
\caption{Composition table derived from the shortest path in the $CNG_{L}$ describing which rotation changes a feature from an initial location/side to another location/side of an object. Commas indicate alternatives and there are blanks between opposite locations since there is no 90 degree rotation that can produce such transformation.}
%Adjacency matrix: how cube sides change their location depending on the rotation applied.}
\label{tab:matrixqualitative}
\begin{adjustbox}{center}
\footnotesize
%\begin{tabularx}{\textwidth}{ |p{2 cm}|p{2 cm}|p{2 cm}|c|c|c|L|  }\hline
%\begin{tabular}{ |p{1cm}| p{2cm} |p{2cm} | p{2cm} | p{2cm} | p{2cm} | p{2cm} | }\hline

\begin{tabular}{| r >{\centering\arraybackslash}m{0.15in} || >{\centering\arraybackslash}m{0.25in} |>{\centering\arraybackslash}m{0.25in} |>{\centering\arraybackslash}m{0.25in} |>{\centering\arraybackslash}m{0.25in} |>{\centering\arraybackslash}m{0.25in} |>{\centering\arraybackslash}m{0.25in} |}\hline
\textcolor{gray}{\textit{to:}}  &front&back&up&down&right&left\\ \hline
\textcolor{gray}{\textit{from:}}  &&&&&&\\ 
front& $\curvearrowleft$, $\curvearrowright$ &  &\textuparrow & \textdownarrow &\textrightarrow &\textleftarrow \\ \hline
back&  & $\curvearrowleft$, $\curvearrowright$  &\textdownarrow &\textuparrow &\textleftarrow &\textrightarrow \\ \hline
up&\textdownarrow &\textuparrow & \textrightarrow, \textleftarrow & &$\curvearrowright$&$\curvearrowleft$\\ \hline
down&\textuparrow &\textdownarrow & & \textrightarrow, \textleftarrow &$\curvearrowleft$&$\curvearrowright$\\ \hline
right&\textleftarrow &\textrightarrow &$\curvearrowleft$&$\curvearrowright$& \textdownarrow,\textuparrow & \\ \hline
left&\textrightarrow &\textleftarrow &$\curvearrowright$&$\curvearrowleft$& &  \textdownarrow,\textuparrow \\  \hline 
\end{tabular}
\end{adjustbox}
\end{table}
%%%%%%%%%%%%%%%%%%%%%%%%%%%%%%%%%%%%%%%%%%%%%%%%%%%%

%%%%%%%%%%%%%%%%%%%%%%%%%%%%%%%%%%
\begin{table}[h!]
\caption{Shortest location changes between visible sides in the CCT (fru).}
\label{tab:MostVisible}
\begin{adjustbox}{center}
\footnotesize
\begin{tabular}{|c|c|c|c|}\hline
        &front & up & right \\ \hline
front & - & \textuparrow & \textrightarrow\\ 
up & \textdownarrow & - & $\curvearrowright$\\ 
right & \textleftarrow & $\curvearrowleft$ & -\\ \hline
\end{tabular}
\end{adjustbox}
\end{table}
%%%%%%%%%%%%%%%%%%%%%%%%%%%%%%%%%%%

From \tablename{} \ref{tab:matrixqualitative}, rotations which change feature locations between visible sides in the CCT (fru) can be extracted. If the inferred rotation produces all the needed location changes in the scene, then we must check if the orientation of such features is still consistent with the inferred rotation (using \tablename{} \ref{tab:ChangeOriVisible}). For example a symbol with the orientation 0q on the \textit{front}, after the movement (\textuparrow) will remain with 0q orientation in the side up (\textit{same} orientation). But if a symbol with orientation 0q is on the side \textit{right} after the movement ($\curvearrowleft$), will be on the side \textit{up} with 3q orientation. So if a symbol goes from \textit{right} to  \textit{up} it will lose 1q of the orientation and if goes from  \textit{up} to \textit{right} it will be added to its orientation 1q, this relation can be seen in \tablename{} \ref{tab:ChangeOriVisible}.

%%%%%%%%%%%%%%%%%%%%%%%%%%%%%%%%%%%%%%%%%%%%
\begin{table}[h!]
\caption{Table of orientation changes between translations on visible sides.}
\label{tab:ChangeOriVisible}
\begin{adjustbox}{center}
\footnotesize
\begin{tabular}{| c |c | c | c }\hline
   & front & up & right\\ \hline
front&  & $same$ & $same$ \\ 
up & $same$ &  & +1q  \\ 
right & $same$ & -1q &   \\ \hline
\end{tabular}
\end{adjustbox}
\end{table}
%%%%%%%%%%%%%%%%%%%%%%%%%%%%%%%%%%%%%%%%%%%%%%

%%%%%%%%%%%%%%%%%%%%%%%%%%%%%%%%%%%

%%%%%%%%%%%%%%%%%%%%%%%%%%%%%%%%%%%%%%%%%%%%%%%%%%%%%%%%%%%%%%%%%%%%%%%%%%%%%%%%%%%%%%%%%%%
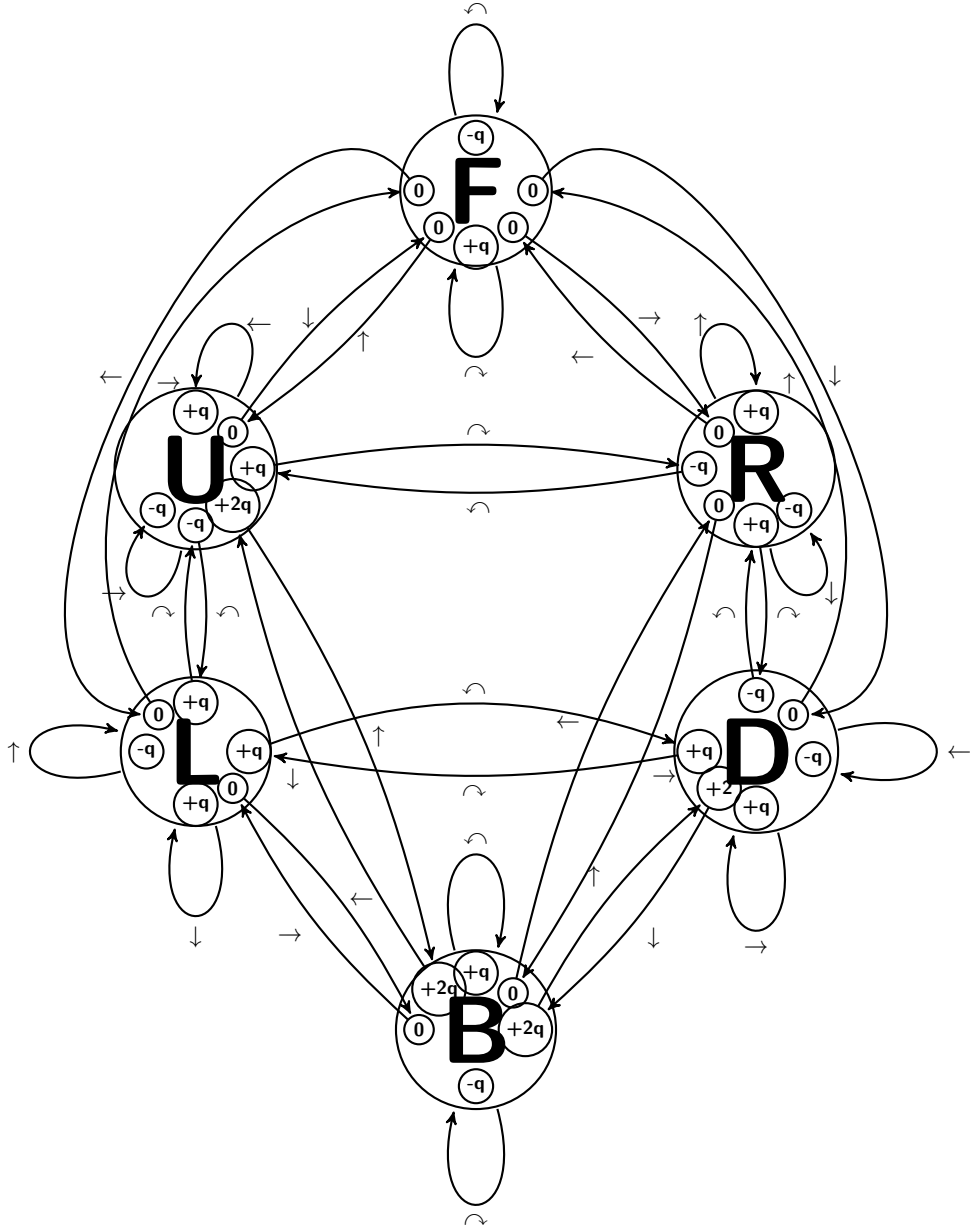
\begin{figure*}[ht]
\caption{Conceptual neighborhood graph relating the Rotation movement to the change of Location and the change of orientations ($CNG_{RLO}$). Note that orientation change is described as an increment: \{0, +q, -1, +2q\}.}
\label{fig:graph-orientation}
\begin{adjustbox}{center}
\begin{tikzpicture}[->,>=stealth',shorten >=1pt,auto,node distance=3.5cm,scale=0.5,
                    thick,main node/.style={circle,draw,scale=0.5,font=\sffamily\Large\bfseries}]

  %big nodes  
  \node[main node,scale=5] (F) {F};
  \node[main node,scale=5] (U) [below left of = F,xshift=1cm,yshift=1cm] {U};
  \node[main node,scale=5] (R) [below right of = F,xshift=-1cm,yshift=1cm] {R};
  \node[main node,scale=5] (L) [below of = U,xshift=0cm,yshift=2cm] {L};
  \node[main node,scale=5] (D) [below of = R,xshift=0cm,yshift=2cm] {D};
  \node[main node,scale=5] (B) [below right of = L,xshift=-1cm,yshift=1cm] {B};

%small nodes
  \node[main node,circle,scale=1] (F1) [below right of=F,xshift=-1.5cm,yshift=1.5cm] {0};%below right
  \node[main node,circle,scale=1] (F2) [below left of=F,xshift=1.5cm,yshift=1.5cm] {0};
  \node[main node,circle,scale=1] (F3) [right of=F,xshift=-2cm,yshift=0cm] {0};
  \node[main node,circle,scale=1] (F4) [left of=F,xshift=2cm,yshift=0cm] {0};
  \node[main node,circle,scale=1] (F5) [above of=F,xshift=0cm,yshift=-2.1cm] {-q};
  \node[main node,circle,scale=1] (F6) [below of=F,xshift=0cm,yshift=2cm] {+q};
  
  \node[main node,circle,scale=1] (U1) [above right of=U,xshift=-1.5cm,yshift=-1.5cm] {0};%below right
  \node[main node,circle,scale=1] (U2) [below right of=U,xshift=-1.5cm,yshift=1.5cm] {+2q};
  \node[main node,circle,scale=1] (U3) [right of=U,xshift=-2cm,yshift=0cm] {+q};
  \node[main node,circle,scale=1] (U4) [below of=U,xshift=0cm,yshift=2cm] {-q};
  \node[main node,circle,scale=1] (U5) [left of=U,xshift=2.5cm,yshift=-1.1cm] {-q};
  \node[main node,circle,scale=1] (U6) [above of=U,xshift=0cm,yshift=-2cm] {+q};
  
  \node[main node,circle,scale=1] (L1) [right of=L,xshift=-2.1cm,yshift=0cm] {+q};%below right
  \node[main node,circle,scale=1] (L2) [above left of=L,xshift=1.5cm,yshift=-1.5cm] {0};
  \node[main node,circle,scale=1] (L3) [below right of=L,xshift=-1.5cm,yshift=1.5cm] {0};
  \node[main node,circle,scale=1] (L4) [above of=L,xshift=0cm,yshift=-2.2cm] {+q};
  \node[main node,circle,scale=1] (L5) [left of=L,xshift=2.2cm,yshift=0cm] {-q};
  \node[main node,circle,scale=1] (L6) [below of=L,xshift=0,yshift=2.1cm] {+q};
  
  \node[main node,circle,scale=1] (B1) [above right of=B,xshift=-1.5cm,yshift=-1.5cm] {0};%below right
  \node[main node,circle,scale=1] (B2) [above left of=B,xshift=1.5cm,yshift=-1.4cm] {+2q};
  \node[main node,circle,scale=1] (B3) [right of=B,xshift=-2.2cm,yshift=0cm] {+2q};
  \node[main node,circle,scale=1] (B4) [left of=B,xshift=2cm,yshift=0cm] {0};
  \node[main node,circle,scale=1] (B5) [above of=B,xshift=0cm,yshift=-2cm] {+q};
  \node[main node,circle,scale=1] (B6) [below of=B,xshift=0cm,yshift=2cm] {-q};
  
  \node[main node,circle,scale=1] (D1) [left of=D,xshift=2cm,yshift=0cm] {+q};%below right
  \node[main node,circle,scale=1] (D2) [above right of=D,xshift=-1.5cm,yshift=-1.5cm] {0};
  \node[main node,circle,scale=1] (D3) [below left of=D,xshift=1.5cm,yshift=1.5cm] {+2};
  \node[main node,circle,scale=1] (D4) [above of=D,xshift=0cm,yshift=-2cm] {-q};
  \node[main node,circle,scale=1] (D5) [right of=D,xshift=-2cm,yshift=-0.2cm] {-q};
  \node[main node,circle,scale=1] (D6) [below of=D,xshift=0cm,yshift=2cm] {+q};
  
  \node[main node,circle,scale=1] (R1) [above left of=R,xshift=1.5cm,yshift=-1.5cm] {0};%below right
  \node[main node,circle,scale=1] (R2) [below left of=R,xshift=1.5cm,yshift=1.5cm] {0};
  \node[main node,circle,scale=1] (R3) [left of=R,xshift=2cm,yshift=0cm] {-q};
  \node[main node,circle,scale=1] (R4) [below of=R,xshift=0cm,yshift=2cm] {+q};
  \node[main node,circle,scale=1] (R5) [right of=R,xshift=-2.5cm,yshift=-1.1cm] {-q};
  \node[main node,circle,scale=1] (R6) [above of=R,xshift=0cm,yshift=-2cm] {+q};

   \path[every node/.style={font=\sffamily\small}]
    (F1) edge [bend left=10] node [right] {\textrightarrow } (R1)
    (F2) edge [bend left=10] node[right] {\textuparrow } (U1)
    (F4) edge [bend right=115] node[left] {\textleftarrow } (L2)
    (F3) edge [bend left=115] node[right] {\textdownarrow } (D2)
    (F) edge [loop above] node {$\curvearrowleft$ } (f)
    	edge [loop below] node {$\curvearrowright$ } (f)

    (U3) edge [bend left=10]  node  {$\curvearrowright$ } (R3)
    (U1) edge [bend left=10] node[left] {\textdownarrow } (F2)
    (U2) edge [bend left=10] node[right] {\textuparrow  } (B2)
    (U4) edge [bend left=10] node[right] {$\curvearrowleft$ } (L4)

     (L3)  edge [bend left=10] node[right] {\textleftarrow } (B4)
     (L2) edge [bend left=60] node[right] {\textrightarrow } (F4)
     (L1)  edge  [bend left=20] node {$\curvearrowleft$ } (D1)
     (L4)  edge [bend left=10] node[left] {$\curvearrowright$ } (U4)
     (L)  edge [loop left] node {\textuparrow } (l) 
     	  edge [loop below] node {\textdownarrow } (l)

   	(B4)   edge  [bend left=10] node  {\textrightarrow } (L3)
    (B2)   edge [bend left=10] node {\textdownarrow } (U2)
    (B1)   edge [bend left=10] node {\textleftarrow } (R2)
    (B3)   edge [bend left=10] node {\textuparrow} (D3)
    (B) edge [loop above] node {$\curvearrowleft$ } (b)
    	edge [loop below] node {$\curvearrowright$ } (b)

  	 (D3)  edge [bend left=10] node  {\textdownarrow } (B3)
     (D4)  edge [bend left=10] node {$\curvearrowleft$ } (R4)
     (D1)  edge [bend left=10] node {$\curvearrowright$ } (L1)
     (D2)  edge [bend right=60] node[left] {\textuparrow  } (F3)
     (D)  edge [loop right] node {\textleftarrow } (D) 
     	  edge [loop below] node {\textrightarrow } (D)

    (R1) edge [bend left=10] node  {\textleftarrow } (F1)
    (R2)  edge [bend left=10] node {\textrightarrow } (B1)
    (R3) edge [bend left=10] node {$\curvearrowleft$ } (U3)
    (R4)  edge [bend left=10] node {$\curvearrowright$ } (D4);
    
    \path (U) edge [out=60,in=90,looseness=6] node[right] {\textleftarrow} (U6);
    \path (U) edge [out=260,in=230,looseness=5] node[left] {\textrightarrow} (U5);
    
     \path (R) edge [out=120,in=90,looseness=6] node[left] {\textuparrow} (R);
      \path (R) edge [out=280,in=310,looseness=5] node[right] {\textdownarrow} (R);

\end{tikzpicture}
\end{adjustbox}
\vspace{-35pt}
\end{figure*}
%%%%%%%%%%%%%%%%%%%%%%%%%%%%%%%%%%%%%%%%%%%%%%%%%%%%%%%%%%%%%%%%%%%%%%%%%%%%%%%%%%%%%%%

\section{Solving the Cube Comparison Test (CCT)}
\label{sec:heuristic}

%As humans do not typically engage in exploring all the problem space \citep{Helie-Pizlo:2021}, in order to solve the Cube Comparison Test (CCT), the QOR model is applied together with an heuristic.
%In the literature, heuristics have shown to provide more effective behaviours in agents, for example the gaze heuristic \citep{Gigerenzer} enabled cognitive agents to effectively catching a falling ball without using any dynamic model of the flying object \citep{Freksa-Kroll:2020}.

This paper proposes that the CCT can be solved by visually comparing the features in both cubes in order to find which features  (i.e. symbols/textures) are repeated in both cubes and how many pairs of repeated features do we have in each scene (pair of cubes). Thus, let us define $R$ as the number of pairs of repeated features, then:

\begin{enumerate}[-]
\item If $R=0$, it indicates that there is no repeated feature. Thus, the features on one cube/object could be the occluded features in the perspective taken on the other cube/object, then both cubes/objects can be the same, but seen from opposite perspectives (i.e. fru--bld; rbu--flu; lfd--rbu, etc).

\item If $R=1$, it indicates that only one feature is common. Then the solution involves to search for a path of rotation actions in the $CNG_{RLO}$ graph that changes the feature location/orientation as observed. And then use this path on the other two features of the original cube to check if they end up on an invisible location on the final cube. If the path of rotations indicates that they must end up on a visible location and they are not showed, then it is not the same cube. Otherwise, it is.

\item if $R=2$, this indicates that two features are in common. Then the solution involves searching for two paths of rotation actions in the $CNG_{RLO}$ graph that produce the observed location/orientation changes, and then compare them to check if they could be the same sequence of rotations. If they are not, the cubes are different. If both sequences of rotations are equivalent, then the path must be applied to the remaining feature to ensure that it ends up on an invisible location.

\item if $R=3$, this indicates that the 3 features in both cubes are repeated. Then the solution is
the same explained for $R=2$ but building 3 paths and compare them. They must contain the same rotation actions so that the cubes can be the same, but seen from different perspectives.
\end{enumerate}

\section{Conclusions and Future Work}
This paper outlines an algorithm to solve the Cube Comparison Test using a Qualitative Object Descriptor (QOD) and Qualitative Object Rotations (QOR) which will be relevant when modelling (i) Human-Computer- Interaction in tasks such as interactive applications developed to train users' spatial skills (e.g. educational videogames or educational applications); and (ii) Human-Robot-Interaction tasks intended to train users' spatial reasoning skills by physical interaction (e.g. building towers  with specific blocks to achieve a specific shape). It will be also relevant for cognitive robotics when autonomous robots/agents need to find out the corresponding rotation to apply to an object so that it has a particular view or when they need to compare an old view of an object stored in memory with a current view of a object in order to find out if both objects are the same or different.

\section*{Acknowledgements}
I acknowledge the funding by the Wallenberg AI, Autonomous Systems and Software Program (WASP) awarded by the Knut and Alice Wallenberg Foundation, Sweden.

%%%%%%%%%%%%%%%%%%%%%%%%%%%%%%%%%%%%%%%%%%%%%%%%%%%%%%%%%%%%%%%%%%%%%%%%%%%%%%%%%%%%%%%
%
%  Bibliografia
%
%%%%%%%%%%%%%%%%%%%%%%%%%%%%%%%%%%%%%%%%%%%%%%%%%%%%%%%%%%%%%%%%%%%%%%%%%%%%%%%%%%%%%%%

\bibliographystyle{apalike}
\bibliography{autosam,Spatial-references}

\end{document}